\documentclass{article}
\usepackage{spconf,amsmath,graphicx}

\usepackage[left=19mm, right=19mm, top=35mm, bottom=25mm]{geometry} 
\usepackage{hyperref}
\usepackage{amssymb, bm, mathrsfs}
\usepackage{xcolor, adjustbox, booktabs, multirow, colortbl, array, caption}

\definecolor{highlight}{RGB}{255, 223, 186} 
\definecolor{tablegray}{HTML}{EFEFEF} 

\title{A Novel Context-adaptive Fusion of Shadow and Highlight Regions for Efficient Sonar Image Classification}

\twoauthors
  {Kamal Basha S, Anukul Kiran B, Athira Nambiar}
  {Department of Computational Intelligence,\\
   SRM Institute of Science and Technology,\\
   Chennai, India\\
   }
  {Suresh Rajendran}
  {Department of Ocean Engineering,\\
   Indian Institute of Technology Madras,\\
   Chennai, India\\}

\begin{document}
\maketitle
\begin{abstract}
Sonar imaging is fundamental to underwater exploration, with critical applications in defense, navigation, and marine research. Shadow regions, in particular, provide essential cues for object detection and classification, yet existing studies primarily focus on highlight-based analysis, leaving shadow-based classification underexplored. To bridge this gap, we propose a \textbf{Context-adaptive sonar image classification framework} that leverages advanced image processing techniques to extract and integrate discriminative shadow and highlight features. Our framework introduces a novel \textbf{shadow-specific classifier} and adaptive \textbf{shadow segmentation}, enabling effective classification based on the dominant region. This approach ensures optimal feature representation, improving robustness against noise and occlusions. In addition, we introduce a \textbf{Region-aware denoising model} that enhances sonar image quality by preserving critical structural details while suppressing noise. This model incorporates an explainability-driven optimization strategy, ensuring that denoising is guided by feature importance, thereby improving interpretability and classification reliability. Furthermore, we present \textbf{S3Simulator+}, an extended dataset incorporating naval mine scenarios with physics-informed noise specifically tailored for the underwater sonar domain, fostering the development of robust AI models. By combining novel classification strategies with an enhanced dataset, our work addresses key challenges in sonar image analysis, contributing to the advancement of autonomous underwater perception.

\end{abstract}
\vspace{-0.3cm}


%
\section{Introduction}
\vspace{-0.3cm}
Sonar imaging plays a crucial role in underwater exploration, navigation, and object detection by capturing acoustic echoes to form images in conditions where optical methods fail~\cite{blondel2010handbook}. However, sonar images suffer from coarse resolution, noise, and acoustic shadows, which pose challenges for accurate analysis. Traditional approaches prioritize highlight regions while often overlooking shadows, despite their rich structural and geometric information. This under-utilization limits classification performance, particularly in side-scan sonar imaging, where shadows can provide more reliable object, especially in the object classes like Mine-like objects (MLO) .

To address these challenges, we introduce a novel \textbf{context-adaptive sonar image classification framework} that integrates shadow and highlight information for robust sonar image analysis. \textcolor{black}{This is one of the first holistic models that adaptively classifies objects by analyzing both global (shadow and highlight) and local (shadow-specific) sonar features. By leveraging contextual information, it enhances classification accuracy and robustness in diverse sonar scenarios.} Our method employs LAB-LCH transformation for feature isolation, spectral ratio computation for shadow enhancement, and noise suppression via average convolution. Shadow regions are precisely refined using K-means clustering and morphological operations. A \textbf{dual-stream architecture}, consisting of separate MLPs for shadow-only and shadow-plus-highlight streams, enables adaptive feature prioritization via an attention mechanism, significantly improving classification accuracy in cluttered underwater environments.

In addition to context-adaptive classification, we propose a \textbf{region-aware denoising model}, designed to preserve critical features in sonar images during denoising. In sonar imaging, the loss of information due to noise removal can have a significant impact on classification performance~\cite{abu2018robust}. Our model ensures that essential features are maintained, preventing information loss that could compromise the integrity of the analysis. This work addresses the challenge of denoising in sonar imaging, where preserving contextual information in noisy environments is crucial for robust object detection.

Further, the scarcity of high-quality sonar datasets hampers AI advancements in this domain. Existing datasets often lack diversity, suffer from confidentiality constraints, or fail to incorporate realistic underwater noise conditions. To address this, we introduce \textbf{S3Simulator+}, an updated version of S3Simulator~\cite{kamal2025s3simulator}, featuring an enhanced synthetic dataset with high-fidelity naval mine classes, advanced shadow rendering, and a range of physics-informed noise profiles, including reverberation, backscatter, and multipath interference.

This work highlights the importance of shadows in sonar-based object classification by integrating a robust, holistic model with explainable denoising. A fusion framework enhances reliability while preserving key features, and a high-fidelity naval mine dataset aids benchmarking and adaptability across underwater scenarios.
%

\vspace{-0.5cm}

\section{Related Work}
\vspace{-0.3cm}
\textbf{Sonar Image Classification}  
Sonar image classification has been explored using various machine learning and deep learning techniques. Random Forest and Linear Regression have been used to distinguish rocks from mines \cite{sowmya2024machine}. Transfer learning has been utilized for sonar image classification, where pre-trained models like VGG-19, ResNet50, and EfficientNet were fine-tuned for sediment classification using side-scan sonar images \cite{chandrashekar2023side}. The ResNet-ACW network integrates few-shot learning and GAN-based data augmentation \cite{preciado2022self}. Additionally, the Sparse Kullback-Leibler Divergence Fuzzy C-Means (SKLFCM) method has been proposed for shadow area detection \cite{si2025unsupervised}.

\textbf{Sonar Datasets}  
Real-world datasets for sonar image classification include the \textit{Seabed Objects-KLSG} dataset \cite{huo2020seabed}, which features 385 wrecks, 36 drowning victims, and various other objects. Similarly, the \textit{AI4Shipwrecks} dataset \cite{sethuraman2021ai4shipwrecks} provides high-resolution side-scan sonar images labeled with marine archaeological expertise. Another dataset, the \textit{Sonar Common Target Detection Dataset (SCTD)} \cite{sctd2020}, contains images of planes, shipwrecks, and drowning victims. Unreal Engine has been used to simulate sonar images \cite{shin2019unreal}, while ray tracing combined with GANs has been employed to produce realistic sonar images \cite{sung2021raygan}. Cycle GANs have been utilized for realistic acoustic dataset generation \cite{liu2021cyclegan}, and diffusion models have been applied for high-quality synthetic sonar image generation \cite{yang2022diffusion}.

Unlike previous studies that primarily focus on highlight-based classification, our work introduces a Context-adaptive fusion framework that integrates shadow-specific features, an explainability-driven denoising model, and an extended sonar dataset. This comprehensive approach enhances classification robustness and interpretability, addressing key limitations in existing sonar image analysis methods.

\vspace{-0.4cm}

\section{Methodology}
\vspace{-0.3cm}
This section outlines the methodology for context-adaptive fusion of shadow and highlight regions in sonar images, consisting of two components: context-adaptive classifier fusion (Section~\ref{Context adaptive}) and the S3Simulator+ dataset development with physics-informed noise (Section~\ref{s3simulator}). These components form the foundation of our framework for enhancing sonar image analysis.
\vspace{-0.45cm}

\subsection{Sonar Image Context-adaptive fusion of shadow and highlight}
\label{Context adaptive}

\begin{figure*}[t]
    \centering
    \includegraphics[width=\textwidth]{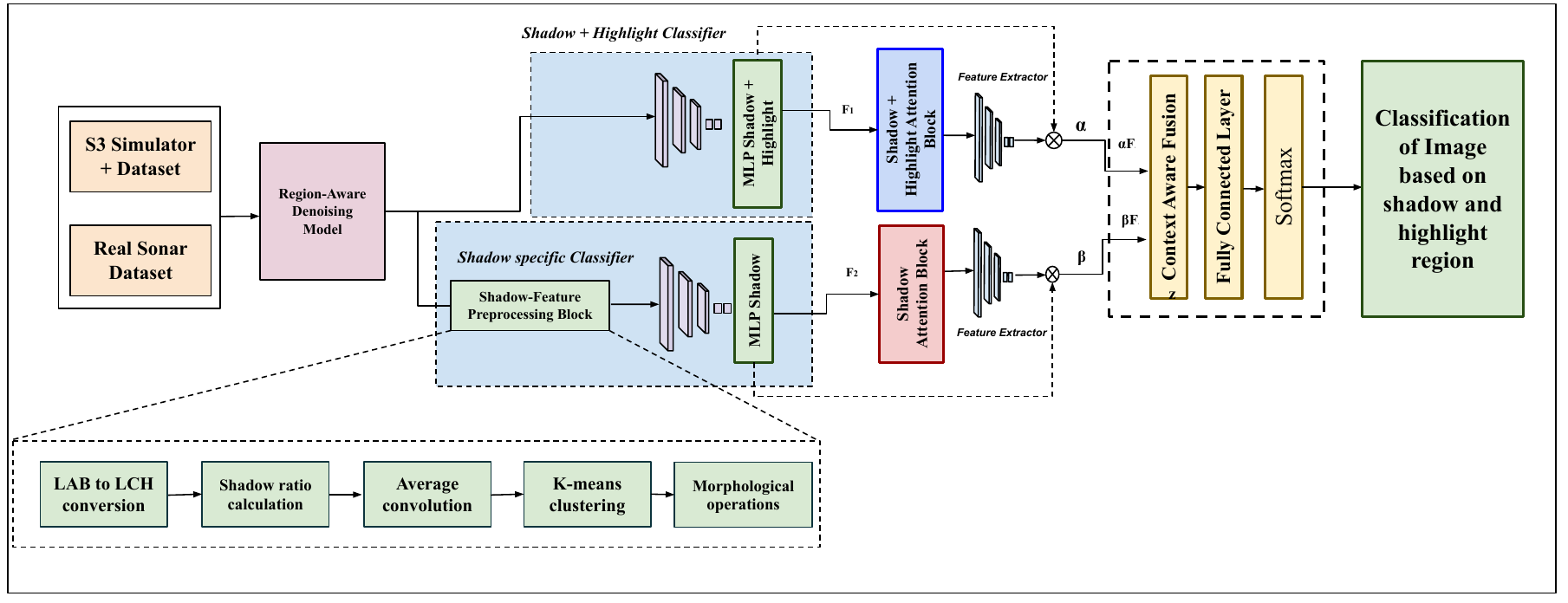}
    \caption{Architecture of Context-adaptive fusion of shadow and highlight regions for efficient sonar image classification}
    \label{fig:archi}
\end{figure*}

To enhance the robustness of shadow-specific feature analysis, both real and S3 Simulator+ sonar images, including noisy and clean datasets, undergo denoising using a novel approach developed to preserve important features in the sonar domain, the Region-aware denoising model which is discussed in Section~\ref{region_aware}. The denoised images are then processed in two parallel streams as shown in the proposed architectural diagram Fig. ~\ref{fig:archi}. In the first stream, they are fed into a classifier with DenseNet\footnote{DenseNet is chosen as the de-facto backbone model after comparative analysis against multiple models i.e. VGG16, Resnet, EfficientNet etc.} as the backbone, capturing a combined representation of shadow and highlight regions for comprehensive analysis. In the second stream, shadow feature preprocessing is performed to extract meaningful shadow-specific information, followed by feature extraction and forwarded to the attention fusion module

\vspace{-0.4cm}
\subsubsection{Shadow feature preprocessing block}
\label{image_preprocessing}
\vspace{-0.2cm}
\textcolor{black}{To achieve accurate shadow segmentation in sonar images, techniques isolate shadow features and enhance contrast between shadow and non-shadow regions.} Methods such as spectral ratio calculation, noise reduction, and clustering improve detection precision while minimizing boundary artifacts.

The preprocessing pipeline in the shadow feature block begins by enhancing shadow-specific features for sonar domain analysis. Sonar images are loaded, remapped to \([H, W, C]\) format, and intensity values are scaled to \([0, 255]\). The RGB channels are retained for analysis, converted to LAB color space, and subsequently mapped to LCH for isolating lightness (\(L\)), chroma (\(C\)), and hue (\(H\)). Normalization of \(L\) and \(H\) ensures consistency across varying illumination. The spectral ratio (\(SR\)) is derived as:
\vspace{-0.2cm}
\begin{equation}
SR = \frac{H + 1}{L + 1},
\end{equation}

\noindent The addition of 1 prevents division by zero, and shadows are emphasized by their lower spectral ratios. A logarithmic transform, $\log(SR + 1)$, is applied to compress the dynamic range. Noise reduction and shadow boundary smoothing are achieved via convolution with a uniform kernel of size \(n \times n\), where kernel values are \(\frac{1}{n^2}\). K-means clustering with \(k+1\) clusters identifies shadow regions, with the threshold determined as minimum Values in each Cluster. Morphological closing, using a disk-shaped structuring element, refines shadow boundaries by sequential dilation and erosion. This two-stream approach ensures that both the global (shadow and highlight regions) and local (shadow-specific features) information are captured effectively via two individual classifiers for robust sonar image analysis.

\begin{figure*}[t]
    \centering
    \includegraphics[width=0.8\textwidth]{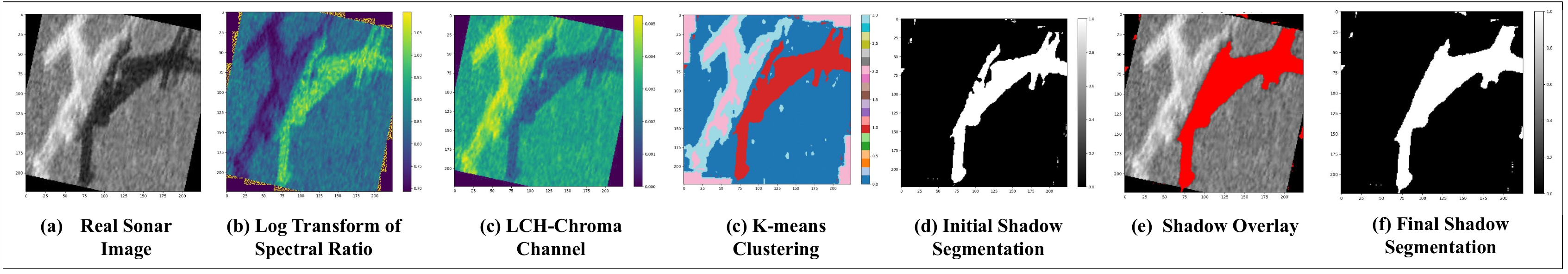}
    \caption{Shadow-Feature Preprocessing Block: Illustration of advanced shadow segmentation techniques utilized in preprocessing to extract meaningful shadow features. (*Additional details are provided in the supplementary material.)}

    \label{fig:image_preprocessing}
\end{figure*}
\vspace{-0.3cm}
\subsubsection{Attention Fusion Model}
\vspace{-0.2cm}
Shadow and highlight information varies by context, affecting the choice between local and global classification. We introduce a context-adaptive selection mechanism, guided by an attention fusion model, to dynamically choose the optimal classifier for robust and accurate sonar image classification.

The attention fusion model combines shadow-aware and raw image features using a dual-stream architecture. Primary features (\(\mathbf{F}_1\)) are extracted from the RGB tensor \(\mathbf{X} \in \mathbb{R}^{B \times C \times H \times W}\) via a backbone network, followed by a multi-layer perceptron (MLP) transformation to \(\mathbf{H}_1\). Shadow-aware features (\(\mathbf{F}_2\)) are processed similarly, yielding \(\mathbf{H}_2\). Global attention weights (\(\bar{\alpha}\) and \(\bar{\beta}\)) are computed for both streams, with normalized weights given as:
\begin{equation}
\alpha = \frac{\|\bar{\alpha}\|}{\|\bar{\alpha}\| + \|\bar{\beta}\|}, \quad \beta = \frac{\|\bar{\beta}\|}{\|\bar{\alpha}\| + \|\bar{\beta}\|}.
\end{equation}

The adaptively fused feature vector is:
\vspace{-0.2cm}
\begin{equation}
\mathbf{Z} = \alpha \mathbf{F}_1 + \beta \mathbf{F}_2.
\end{equation}
The fused feature vector \(\mathbf{Z}\) is passed through a fully connected (FC) layer and softmax to yield class probabilities:
\vspace{-0.2cm}
\begin{equation}
\mathbf{R} = \text{softmax}(\mathbf{W} \cdot \mathbf{Z} + \mathbf{b}),
\end{equation}
where \(\mathbf{W}\) and \(\mathbf{b}\) are trainable parameters. The predicted class is determined by taking the argmax of \(\mathbf{R}\).
\vspace{-0.3cm}

\subsubsection{Objective Function}
\vspace{-0.1cm}
The total loss function is:
\vspace{-0.3cm}
\begin{equation}
L_{\text{total}} = L_{\text{CE}} + \lambda L_{\text{attention}},
\end{equation}
\vspace{-0.2cm}
where \( L_{\text{CE}} \) is the categorical cross-entropy loss, defined as:
\begin{equation}
L_{\text{CE}} = - \sum_{i} t_i \log \text{softmax}(w_k^T \mathbf{Z})_i,
\end{equation}
where \( t_i \) represents the ground truth label for class \( i \), \( w_k \) is the weight vector for class \( k \), and \( \mathbf{Z} \) denotes the extracted feature representation.

The attention regularization term is given by:
\begin{equation}
L_{\text{attention}} = -\sum_{i} w_i \log w_i,
\end{equation}
where \( w_i \) represents the attention weight associated with feature \( i \), encouraging a more uniform distribution of attention across features. The model is optimized using the Adam optimizer with a learning rate of \(0.001\), and \textcolor{black}{ \( \lambda \) is a learnable hyperparameter that balances classification accuracy and attention smoothness, tuned through experimentation.}
\vspace{-0.4cm}

\subsubsection{Region-aware denoising framework}
\label{region_aware}
\vspace{-0.2cm}
\begin{figure}[t]
    \centering
    \label{archi}
    \includegraphics[width=\columnwidth]{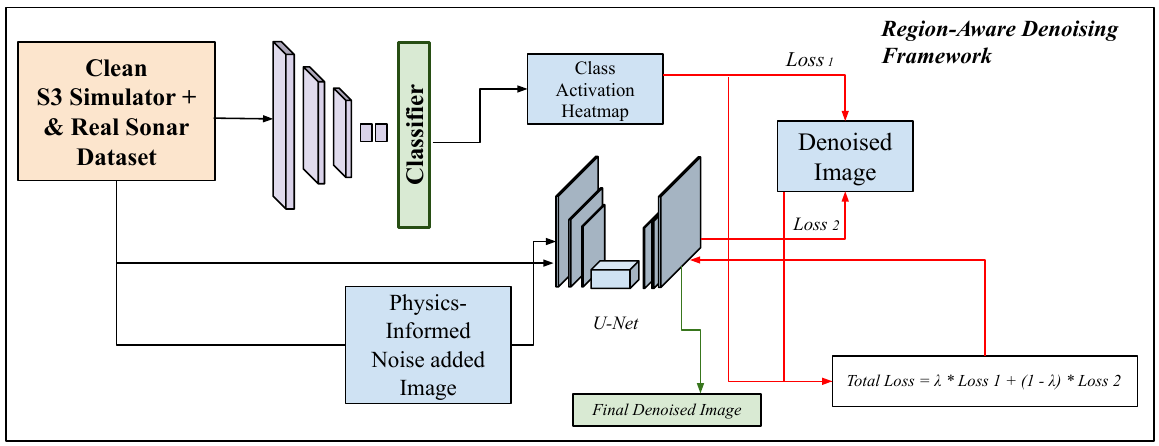}
    \caption{Region aware denoising model}
    \label{fig:visualization}
\end{figure}
Sonar images contain both critical structural details and noise, requiring a denoising approach that selectively enhances important regions. To achieve this, our Region-aware denoising model leverages a U-Net-based architecture\footnote{Inspired by explainable medical image denoising approaches~\cite{dong2023medical}}.
enhanced with Squeeze-and-Excitation (SE) blocks to adaptively refine sonar image denoising. Classical denoising algorithms often risk suppressing key features along with noise, which can degrade classification performance. To mitigate this, we incorporate a feature-preserving loss function motivated by gradient-based eXplainable Artificial Intelligence (XAI), ensuring that essential sonar features remain intact. Given an input noisy sonar image \( X \), the encoder extracts spatial features through convolutional layers, producing a latent representation \( Z \). The SE block applies global average pooling to capture channel-wise feature responses. A two-layer fully connected (FC) network followed by a sigmoid activation generates a set of adaptive weights \( \alpha_c \), which modulate the feature maps:
\vspace{-0.5cm}
\begin{equation}
\alpha_c = \sigma(W_2 \cdot \text{ReLU}(W_1 \cdot s_c))
\end{equation}

where \( W_1, W_2 \) are learned parameters. This enhances \textbf{region-specific denoising} by suppressing irrelevant noise and amplifying informative features. The decoder then reconstructs the denoised image \( \hat{X} \).

The \textbf{custom loss function} integrates three objectives: \textbf{Mean Squared Error (MSE)} to minimize pixel-wise differences between \( \hat{X} \) and the real sonar image \( X_r \), \textbf{Perceptual Loss}~\cite{johnson2016perceptual} to preserve high-level image structures, and \textbf{Region-aware masked loss}, which applies a Grad-CAM mask~\cite{selvaraju2017grad} \( M \) to focus on critical sonar regions. The loss function is formulated as:
\vspace{-0.6cm}

\begin{equation}
\mathcal{L} = \lambda_1 \| \hat{X} - X_r \|^2 + \lambda_2 \| (\hat{X} - X_r) \cdot (1 - M) \|^2
\end{equation}

where \( \lambda_1 \) and \( \lambda_2 \) are weighting factors. This formulation ensures that denoising is both \textbf{perceptually aware} and \textbf{region-adaptive}, improving noise suppression while preserving structural details in sonar images.

\vspace{-0.5cm}
\subsection{Benchmark Dataset Creation}
\label{s3simulator}
\vspace{-0.2cm}
The creation of the S3Simulator+ dataset consists of 7,000 to 8,000 images of ship, plane, and mine classes, beginning with the modeling of real naval mines using AutoCAD Fusion~\cite{autodesk_fusion360}. Three distinct naval mine geometries—truncated cone, cylindrical, and spherical—are developed to represent a diverse range of underwater objects, as shown in Fig.~\ref{fig:s3sim_stages}. Additional details on the dataset generation process can be found in the supplementary material. These 3D models are deployed in the Gazebo~\cite{gazebo_simulator} simulation environment to render synthetic sonar images under varying conditions, considering object positions, orientations, and shadow effects. Computational imaging techniques replicate sonar appearances, generating nadir zones, color-mapped, and grayscale images for different analyses. Additionally, physics-informed noise is incorporated to enhance realism, simulating real-world sonar disturbances such as multipath interference, backscatter and reverberation as detailed in Table~\ref{tab:noise_details}. The dataset is designed to support the development of robust AI models for sonar-based object recognition and classification.  



\begin{table}[ht]
\footnotesize
\centering
\caption{Physics-Informed Noise Details}
\label{tab:noise_details}
\begin{tabular}{|p{2cm}|p{6cm}|}
\hline
\textbf{Noise} & \textbf{Details} \\ \hline
\textbf{Multipath Interference Noise} & Occurs when sonar signals reflect off surfaces, creating multiple signal paths that arrive at different times~\cite{feng2025tdoa}. \\ \hline
\textbf{Backscatter Noise} & Caused by scattering of sonar signals by particles such as sand, plankton, or suspended matter~\cite{quinn2005backscatter}. \\ \hline
\textbf{Reverberation Noise} & Results from multiple reflections of sonar waves in underwater environments, especially from the seabed and surface~\cite{BJORNO2017297}. \\ \hline
\end{tabular}
\end{table}

\vspace{-.8cm}

\section{Experiment Results}
\label{sec:Experiment Results}
\label{sec:Experiment Results}
\vspace{-0.2cm}

\subsection{Quantitative Results}
\label{quantitative}
\vspace{-0.2cm}
This section evaluates the performance of different classifiers within the context-adaptive fusion framework using real sonar images. The experiment leverages transfer learning with DenseNet121 as the backbone and incorporates an adaptive denoising model into the pipeline. It can be observed from the analysis chart  as shown in Table~\ref{tab:classifier_accuracy} that \textcolor{black}{our context-adaptive fusion classifier outperforms the other models, achieving 96.75\% accuracy, compared to individual Shadow + Highlight classifier (92.18\%) and Shadow classifier (85.98\%).}

\begin{table}[ht]
\centering
\begin{tabular}{|l|c|}
\hline
\textbf{Classifier} & \textbf{Accuracy} \\
\hline
Shadow + Highlight Classifier & 92.18\% \\
Shadow Classifier & 85.98\% \\
Context-adaptive Fusion & 96.75\%\\
\hline
\end{tabular}
\caption{Classification accuracy of different classifiers.}
\label{tab:classifier_accuracy}
\end{table}

\begin{figure}[!t]
    \centering
    \includegraphics[width=\linewidth]{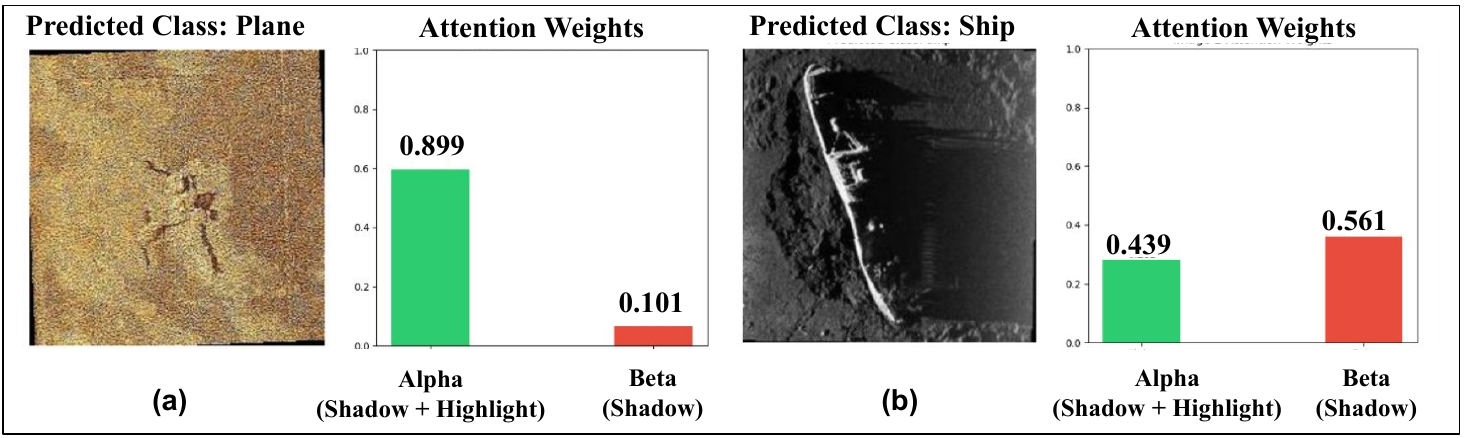}
    \caption{Illustration showing the input of attention weights alpha and beta at various contexts}
    \label{fig:quant_noise}
\end{figure}

\textcolor{black}{The intuitive visualization of attention weights for shadow and shadow + highlight as illustrated in Fig.~\ref{fig:quant_noise} depicts the impact of alpha and beta values in various scenarios. In Fig. 4(a), an image classified as ``Plane" has a dominant alpha (Shadow + Highlight) weight (0.899) over beta (Shadow) (0.101), indicating reliance on combined features. Conversely, Fig. 4(b) shows a ``Ship" classification with a higher beta weight (0.561) than alpha (0.439), emphasizing the importance of shadow features. These results accentuate the model's ability to dynamically \textit{adjust focus based on context} for accurate sonar image classification.} 

\vspace{-0.4cm}
\subsection{Qualitative Results}
\label{Qualitative}
\vspace{-0.2cm}
\subsubsection{S3Simulator+ Image Generation}
\begin{figure}[!t]
    \centering
    \includegraphics[width=\linewidth]{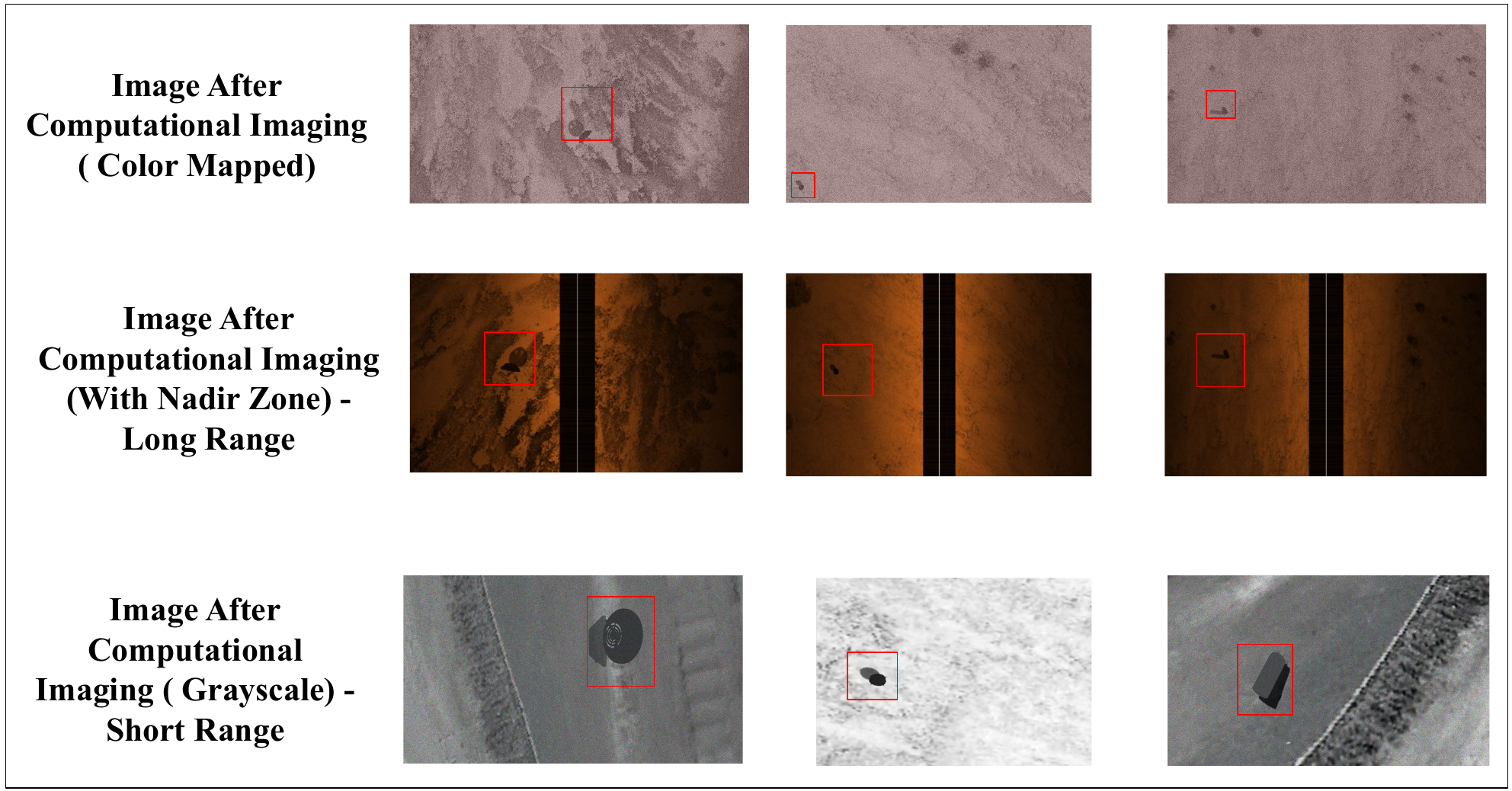}
    \caption{Sample Images from S3Simulator + dataset of naval mines with colormapped, nadir zone in short and long range.}
    \label{fig:s3sim_stages}
\end{figure}

\vspace{-0.2cm}
To demonstrate the progression of mine representation and the diversity of the dataset, Fig.~\ref{fig:s3sim_stages} showcases stages of mine generation and processing. It highlights variations such as mines with nadir zones, as well as close-range (high-detail) and long-range (lower-resolution) sonar images, captured in diverse conditions. (*More results in Supplementary)

\vspace{-0.3cm}
\subsubsection{Physics-Informed Noise Visualization}
\begin{figure}[!t]
    \centering
    \includegraphics[width=\linewidth]{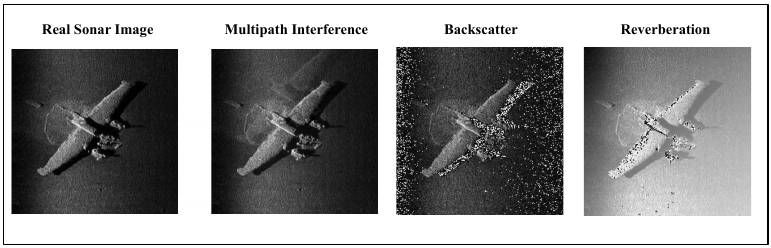}
    \caption{Visualization of physics-informed noise types applied to sonar images, illustrating realistic noise patterns inspired by physical properties of underwater environments. }
    \label{fig:physics_noise}
\end{figure}

\vspace{-0.2cm}
\textcolor{black}{The impact of physics-informed noise (reverberation, multipath interference, and backscatter) on sonar images}, highlighting the realism and domain-specific challenges addressed by our dataset is shown in Fig.~\ref{fig:physics_noise}. This visualization demonstrates how physics-informed noise contributes to domain-specific diversity and enables the development of AI models better suited for challenging underwater environments.

\vspace{-0.4cm}

\subsection{Region aware denoising Visualization}
\begin{figure}[!t]
    \centering
    \includegraphics[width=\linewidth]{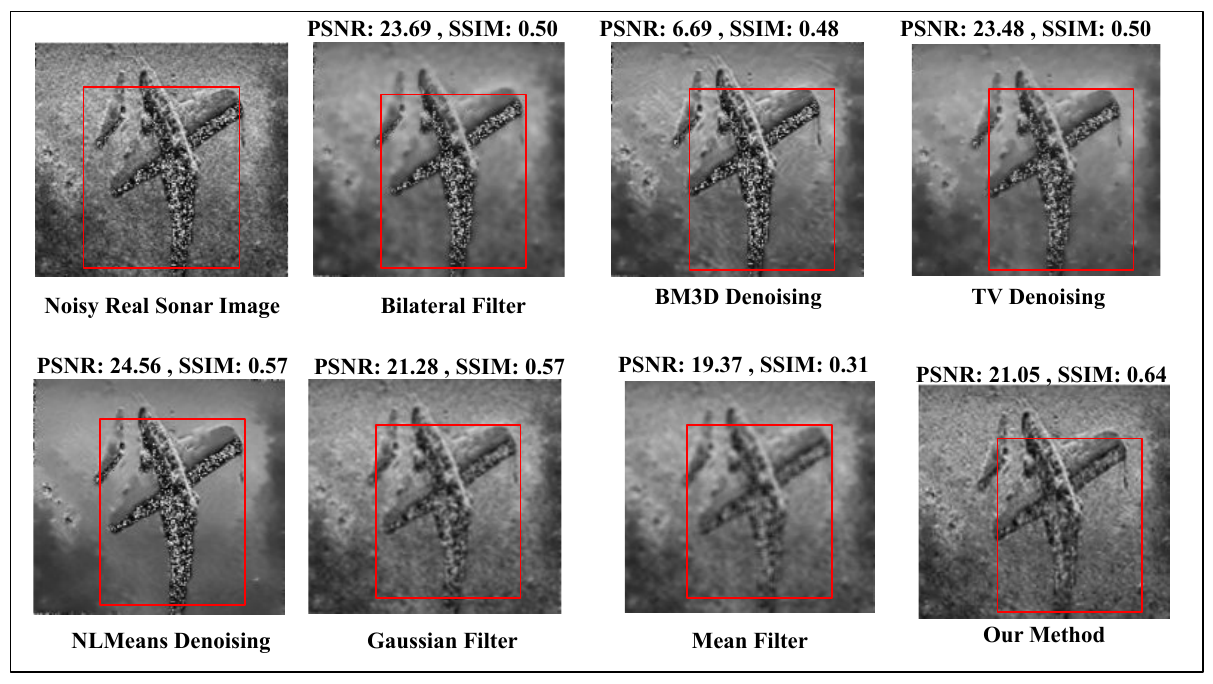}
    \caption{Comparison of denoising methods on a noisy sonar image with visual results and metrics (PSNR, SSIM).}

    \label{fig:denoising_stages}
\end{figure}

\vspace{-0.2cm}
\textcolor{black}{The comparative anaysis of denoising methods on noisy sonar images with our method is depicted in Fig.~\ref{fig:denoising_stages}.} Our method achieves the highest SSIM (0.64), indicating superior structural preservation, and a moderate PSNR (21.05), balancing noise removal and detail retention. The red bounding box highlights clearer restored structures, particularly in shadow regions near high-intensity areas where other models struggle.

\vspace{-0.5cm}
\section{Conclusion}
\vspace{-0.3cm}
We propose a novel approach to sonar image analysis, emphasizing the critical role of shadow regions in classification. \textcolor{black}{Our \textbf{context-adaptive fusion framework for sonar image classification} enhances performance by integrating shadow and highlight features, adapting to the context and depending on the varying environment.} while our region-aware denoising model preserves structural details with explainability-driven optimization. The S3Simulator+ dataset, incorporating naval mine scenarios with physics-informed noise, provides a robust foundation for AI models in real-world underwater conditions. This work advances autonomous underwater perception, improving the robustness and reliability of sonar image analysis.

\bibliographystyle{IEEEbib}
\bibliography{refs}

\end{document}